\documentclass[conference]{IEEEtran}
\usepackage{cite}
\usepackage{amsmath,amssymb,amsfonts}
\usepackage{algorithmic}
\usepackage{graphicx}
\usepackage{textcomp}
\usepackage{xcolor}

\usepackage{commath}
\usepackage{fancyhdr}
\usepackage{yhmath}
\usepackage[ruled,vlined]{algorithm2e}

\def\BibTeX{{\rm B\kern-.05em{\sc i\kern-.025em b}\kern-.08em
    T\kern-.1667em\lower.7ex\hbox{E}\kern-.125emX}}
\begin{document}

\title{Vision-Language Modeling with Regularized Spatial Transformer Networks for All Weather Crosswind Landing of Aircraft\\}

\author{
\textit{Debabrata Pal\textsuperscript{1}, Anvita Singh\textsuperscript{2}, Saumya Saumya\textsuperscript{1}, Shouvik Das\textsuperscript{1}}\\\\
Aerospace, Honeywell Technology Solutions\\
\textsuperscript{1}Bengaluru, Karnataka 560103, India, \textsuperscript{2}Hyderabad, Telangana 500032, India\\
{\{Debabrata.Pal, Anvita.Singh, Saumya.Saumya, Shouvik.Das\}@honeywell.com}\\
}

\maketitle

\begin{abstract}
The intrinsic capability of the Human Vision System (HVS) to perceive depth of field and failure of Instrument Landing Systems (ILS) stimulates a pilot to perform a vision-based manual landing over an autoland approach. However, harsh weather creates challenges, and a pilot must have a clear view of runway elements before the minimum decision altitude. To aid in manual landing, a vision-based system trained to clear weather-induced visual degradations requires a robust landing dataset under various climatic conditions. Nevertheless, to acquire a dataset, flying an aircraft in dangerous weather impacts safety. Also, this system fails to generate reliable warnings, as localization of runway elements suffers from projective distortion while landing at crosswind. To combat, we propose to synthesize harsh weather landing images by training a prompt-based climatic diffusion network. Also, we optimize a weather distillation model using a novel diffusion-distillation loss to learn to clear these visual degradations. Precisely, the distillation model learns an inverse relationship with the diffusion network. Inference time, pre-trained distillation network directly clears weather-impacted onboard camera images, which can be further projected to display devices for improved visibility. Then, to tackle crosswind landing, a novel \textit{Regularized Spatial Transformer Networks} \textsc{(RuSTaN)} module accurately warps landing images. It minimizes the localization error of runway object detector and helps generate reliable internal software warnings. Finally, we curated an aircraft landing dataset (AIRLAD) by simulating a landing scenario under various weather degradations and experimentally validated our contributions.
\end{abstract}
\begin{IEEEkeywords}
Visual landing, runway detection, diffusion, spatial transformer networks, aircraft landing dataset
\end{IEEEkeywords}

\begin{figure}[htbp]
\centerline{\includegraphics[width=\columnwidth]{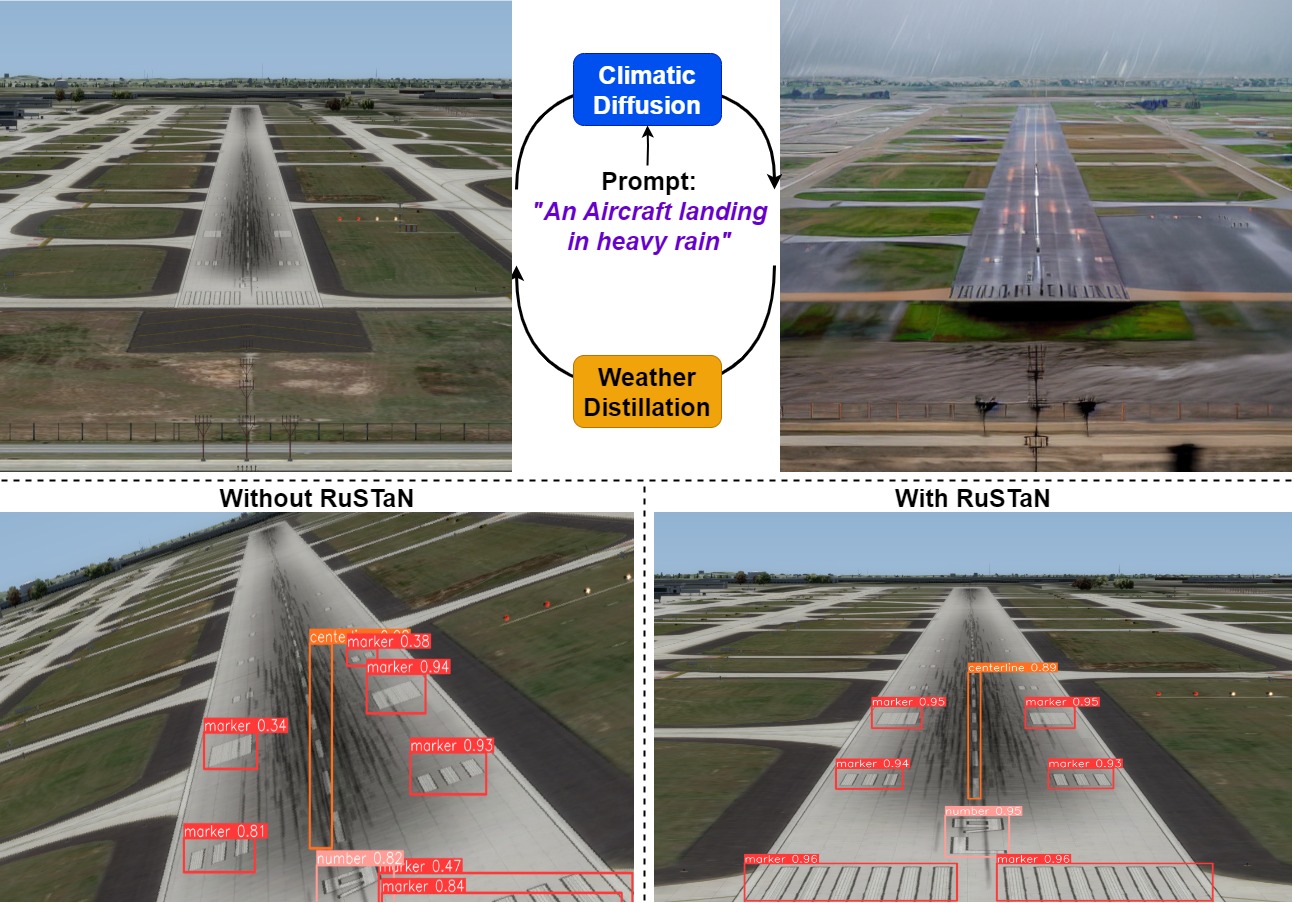}}
\caption{\label{fig:intro} Row-1) During training, we generate harsh weather landing images using a climatic diffusion network, and a distillation module learns to remove those visual degradation. During real-time cockpit onboard inference, the distillation module directly generates clear weather landing images to display to a pilot. Row-2) In case of aircraft rolling, crosswind landing, the \textsc{RuSTaN} module predicts accurate affine parameters to warp an image with a vertical axis parallel to the runway. It helps predict bounding boxes accurately and avoid any missed or false alert in warning generation.}
\end{figure}

\section{Introduction}
Vision-based aircraft landing utilizes visual cues from aircraft camera images to assist a pilot during the landing phase. It acts as a reliable alternative to ILS and is essential for emergency landing at unprepared airfields with no ground-based navigation system. By providing real-time images of the runway and surroundings, visual landing helps pilots gain enhanced situational awareness, especially while landing on short, unknown runways and in low visibility, such as fog, rain, or snow. 

In extremely harsh weather, visibility degrades radically, challenging the usage of visible spectrum as landing assistance. Even though an onboard radar can provide excellent ranging data, the radar images lack spatial resolution and can not provide important runway semantic features like markings and numbers. \textit{It motivates us to propose a human-friendly, visible spectrum-based visual landing system to display an aircraft's exterior environment while removing bad weather-induced visual degradation from real-time camera data.}

Recently, leveraging the deep-learning-based Generative Adversarial Networks (GAN) \cite{radford2015unsupervised}, Weather GAN \cite{li2021weather} and other weather translation models \cite{hwang2022weathergan} have attempted to transfer the weather conditions of on-ground captured weather images from one class to another, such as sunny-to-rainy, sunny-to-snowy. Essentially, they exploited the deep semantic relationship of familiar cues, like blue sky, cloud, wet ground. Nevertheless, a few potential drawbacks exist in directly adapting these weather translation models. Firstly, the distance between a runway and the point where an aircraft prepares for descending is significantly high, allowing the weather to play a dominant role in obscuring the clear runway visibility. Secondly, training these models requires a robust weather-conditioned pixel-paired dataset where it is impossible to repeatedly fly an aircraft through the same geographical coordinates to acquire a real dataset, i.e., capturing the same runway images from the same viewing angle, velocity, altitude, latitude, and longitude during rain, fog, sunlight, etc. As an alternative, CycleGAN \cite{zhu2017unpaired} allows the translation of visual artifacts by learning from unpaired data, but it requires a large-scale landing dataset under different weather categories for training. This is often impractical in our aerial data acquisition context, and also, the CycleGAN predicted images lag granular diversity of mixed weather degradations. This motivates us to generate different weather-conditioned images from simple textual prompts. In essence, using simple English adjectives like `low,' `medium,' and `high,' it is often easy to control different weather conditions in curating a mixed weather degraded dataset.  

The recent emergence of Vision Language Models (VLM) based on stable diffusion \cite{rombach2022high} can generate photo-realistic images from simple textual prompts. Besides, it has been empirically found that the performance of diffusion models is superior to the GAN variants in image synthesis \cite{dhariwal2021diffusion}. Further, ControlNet \cite{zhang2023adding} allows the image generation to be controlled by additional depth maps, edge or pose images. Once the diffusion model generates semantic consistent degraded weather images, we can obtain a clean and harsh weather pixel-paired dataset of the same landing scene without further flying an aircraft in dangerous weather to acquire a dataset. Then, a disjoint model can learn to distill weather artifacts from harsh weather counterparts (Fig. \ref{fig:intro}: Row-1). However, this disentangled diffusion of weather artifacts and filtering by a separate distillation model causes the poor generation of clear weather landing images. The optimization could be optimal with an inverse mapping. It triggers the first research question,\textit{ ``Can we constrain an optimization objective to ensure that the weather distillation model learns an inverse relationship with the harsh weather diffusion network?''}

Once we can generate clear weather landing scenarios, a runway object detector can recognize runway elements and aid the internal software warning generation process, i.e., generating an aural warning, display alerts, etc. However, due to the crosswind landing and flight control, i.e., roll, pitch, yaw, a high offset occurs in the predicted location from reality and introduces false or missed alerts. It motivates us \textit{``Can we perform a real-time accurate projective transformation under crosswind landing scenario, that can aid generating reliable warnings ?''} (Fig. \ref{fig:intro}: Row-2). Notably, the pilot still sees the unwarped but synthetic clear-weather landing images by the distillation module while perfectly getting warned due to the accurate runway object localization of the internally generated warped images by the \textsc{RuSTaN} module.

Finally, we notice that \textit{``the unavailability of a benchmark aircraft landing dataset under degraded weather impedes research on visual landing.''}\\
\noindent\textbf{Our contributions:} We propose a diffusion-distillation loss to guide the weather distillation model in learning the inverse relationship with the climatic diffusion model. Then, to learn how to land at crosswind, we introduce a Regularized Spatial Transformer Networks (\textsc{RuSTaN}) module, which directly optimizes existing STN's \cite{jaderberg2015spatial} Localization Net with randomly sampled known affine parameters. During inference, \textsc{RuSTaN} acts as a regular STN but performs superior in predicting the optimal affine parameters. Finally, we introduce a simulation-based AIRport LAnding Dataset (AIRLAD) under diverse weather scenarios at the busiest airport, Hartsfield-Jackson Atlanta International Airport, to elevate the research on visual landing. The simulation-based landing scenarios are more realistic than the computer graphics-based approach, especially in generating spatiotemporal fog, rain patterns, and noise artifacts like tire skid marks, faded runway markings, crossing taxiways, etc. Further, we leverage a VLM to compose granular complex weather mixtures using simple textual prompts. It significantly reduces paired data formation time. To summarize,
\begin{itemize}
    \item We formulated a diffusion-distillation loss to optimize the climatic diffusion network and weather distillation model jointly.
    \item Proposed \textsc{RuSTaN} precisely warps realtime crosswind landing images, making runway object detection and alert generation reliable.
    \item We curated an aircraft landing image dataset under different weather degradation categories and augmented it with diffusion. 
\end{itemize}

\begin{figure*}[htbp]
\centerline{\includegraphics[width=0.92\linewidth]{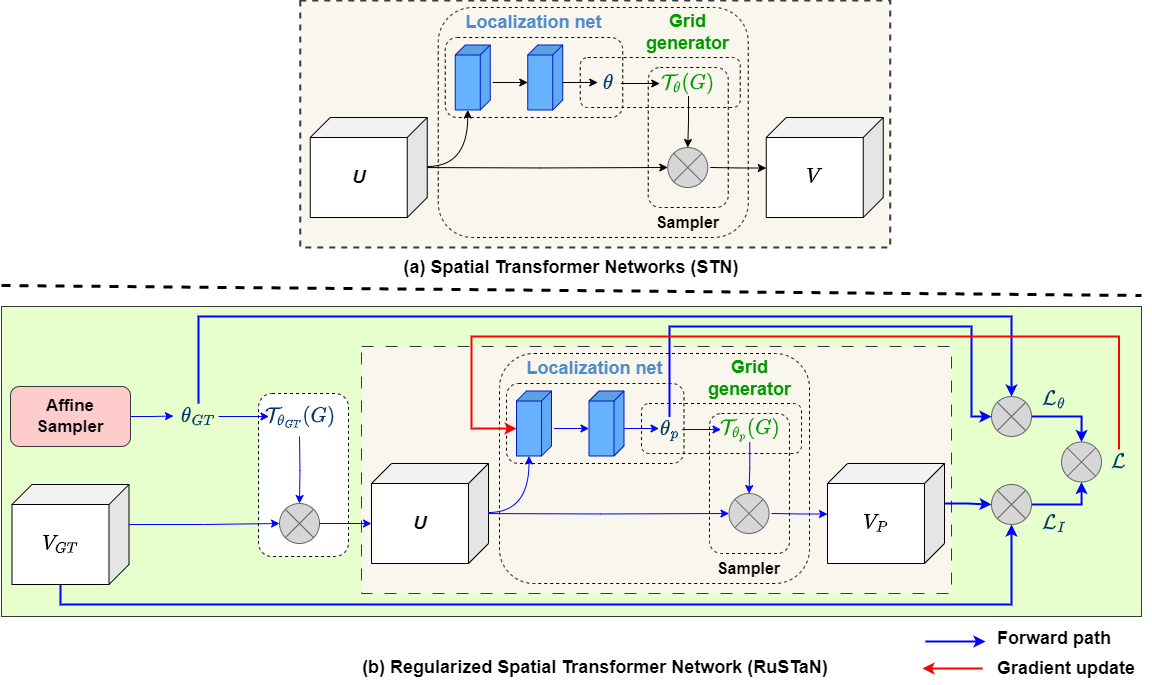}}
\caption{\label{Fig_Rustan} a) In STN, a Localization Net predicts the transformation parameter $\theta$, and an input image or a feature map is warped based on $\mathcal{T}_{\theta}(G)$. Due to the vanishing or exploding gradient in deep networks and optimizing for overall model objective, $\theta$ can be inaccurate. Hence, in b), we regularize Localization Net to predict transformation parameter $\theta_p$ as $\theta_{GT}^{-1}$ during training, where $\theta_{GT}$ is obtained by a novel affine sampler. During inference, \textsc{RuSTaN} acts as STN where Localization Net directly processes real-time projective distorted image $U$ to predict an accurate $\theta$ based on self-supervised pre-training with inversion constraint.}
\label{fig:fig5}
\end{figure*}

\section{Related Work}
\noindent \textbf{i) Image diffusion:} Image diffusion was first conceptualized by Sohl-Dickstein, Jascha, \textit{et al.} \cite{sohl2015deep} and has since gained significant attention in image generation \cite{croitoru2023diffusion, dhariwal2021diffusion, kingma2021variational}. Latent Diffusion Models (LDM) aim to reduce the computational cost by executing diffusion steps within the latent image space \cite{rombach2022high}. Contemporary text-to-image diffusion models leverage pre-trained language models such as CLIP \cite{radford2021learning} to encode textual inputs. Noteworthy among these is Glide \cite{nichol2021glide}, a text-guided diffusion model renowned for its capabilities in image creation. Stable diffusion \cite{rombach2022high} represents a notable large-scale implementation of latent diffusion. In contrast, Imagen \cite{saharia2022photorealistic} diffuses pixels directly through a pyramid structure, bypassing latent image transformations. Despite the enormous success of diffusion models, including its commercial versions like DALL-E2 \cite{ramesh2022hierarchical} and Midjourney \cite{Midjourney}, the generated images still lack context, like the runway centerline is not in the center, uneven distribution 
of runway markers, implausible airport environment, etc. \\ 
\noindent \textbf{ii) Spatial Transformation Networks:} STN \cite{jaderberg2015spatial} allows neural networks to learn spatial transformations of input data, making them invariant to various transformations. STN has been incorporated with GAN, enabling geometric transformations for realistic image composting \cite{lin2018st}. Spectral-spatial transformer networks in remote sensing is designed to classify hyperspectral images by combining spectral and spatial information \cite{zhong2021spectral}. Bas \textit{et al.} \cite{bas20173d} demonstrated the application of 3D morphable models as STNs for efficient 3D face modeling and alignment. Additionally, smart city and urban planning often utilize spatial-temporal transformer networks to forecast traffic flow through spatial dependencies and temporal patterns \cite{xu2020spatial}. However, during training, STN parameters are optimized due to a gradient flow from the overall model objective without any direct calibration against the ground truth affine parameters, which actually causes inaccurate geometric rectification. \\
\noindent \textbf{iii) Runway object detection:} The traditional computer-vision-based runway detection employs HOG descriptors \cite{andreev2019objects} due to photometric invariance but fails to deal with orientation. Damian \textit{et al.} \cite{kordos2023vision} apply a series of segmentation, contour, and line filtering steps with several rules per frame to measure relative aircraft position to the runway, but its computational complexity impedes real-time deployment. To generalize, an Airport Runway Detection Network (ARDN) \cite{amit2021robust} employs a region-based convolutional neural network (R-CNN) but can not deal with harsh weather impacts. LaneNet \cite{wang2018lanenet} and PointLaneNet \cite{chen2019pointlanenet} detect lane markings for automobiles but can not be adapted to the aviation industry due to huge weather impacts at a prolonged distance from the runway. Overall, no runway detection research exists under dynamic weather conditions with versatile projective distortions during the approach stage.

\section{Proposed Method}
\subsection{Problem Statement}
\begin{figure*}[htbp]
\centerline{\includegraphics[width=\linewidth]{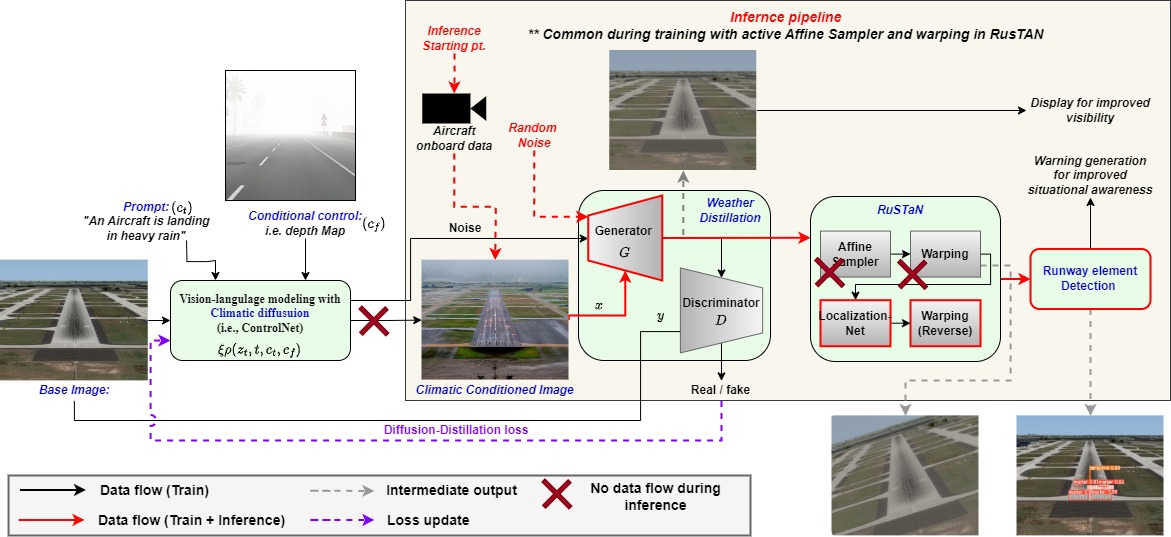}}
\caption{Illustration of the proposed framework. During training, diverse harsh climatic conditions are synthesized using a VLM or climatic diffusion model, and a generative weather distillation module learns to remove the synthesized weather artifacts. Besides, to learn to recognize runway elements during landing at crosswind, the affine sampler in the \textsc{RuSTaN} module geometrically distorts an input image and, followed by a Localization Net, learns to restore from the projective transformation. Finally, an object detector localizes essential runway elements accurately to generate reliable alerts and aid the pilot in a vision-based landing. Inference time data flow is highlighted in `Red' color, which relaxes VLM and affine sampler usage as weather degradation and projective distortion at crosswind can inherently impact the real-time camera data.}
\label{fig:Archi}
\end{figure*}

We address the problem of recognizing essential runway elements under harsh weather conditions for monocular vision-based safe aircraft docking. Landing in extreme weather impacts clear visibility. Also, a non-linear projective transformation distorts the onboard camera data, especially while docking at a heavy crosswind, which must be calibrated first for the reliable localization of the runway elements. In \textit{section B}, we analyze the contemporary approach and find its drawbacks in restoring projective distortion. Then, the newly proposed regularization technique is elaborated in \textit{section C}. Finally, we describe the end-to-end visual aircraft landing pipeline (Fig. \ref{fig:Archi}) with a novel climatic synthesis method in \textit{section D}.

\subsection{Review of the Spatial Transformer Networks:}
STN \cite{jaderberg2015spatial} consists of a Localization Net, a grid generator and is followed by a sampler, as shown in Fig. \ref{Fig_Rustan}a. Incorporating STN in a CNN model, we gain spatial invariance to different geometric variations such as translation, rotation, scaling, and non-rigid skew deformation, etc. Basically, an STN module in a CNN classifier transforms an input image to a canonical shape and adapts its parameters by backpropagation from the final classification layer. The Localization Net predicts the affine parameters, and the grid generator with a sampler performs bilinear transformation. For a 2D affine transformation, the affine matrix ($\mathcal{A}_{\theta}$) can be defined,
\begin{small}
\begin{equation}
\begin{pmatrix}
x_i^s \\
y_i^s 
\end{pmatrix}
= \mathcal{T}_{\theta}(G) =
\mathcal{A}_{\theta} 
\begin{pmatrix}
x_i^t \\
y_i^t \\
1
\end{pmatrix}
=
\begin{bmatrix}
\theta_{11} & \theta_{12} & \theta_{13}\\
\theta_{21} & \theta_{22} & \theta_{23}
\end{bmatrix}
\begin{pmatrix}
x_i^t \\
y_i^t \\
1
\end{pmatrix}
\end{equation}
\end{small}

Where, an input image $U$ is warped by an affine transformed regular grid $G$, $\mathcal{T}_{\theta}(G)$, to produce an output image $V$. $(x_i^s, y_i^s)$ and $(x_i^t, y_i^t)$ are the normalized regular grid $G$ coordinates of $U$ and $V$ with spatial bounds as $[-1,1]$.

However, we observe that STN suffers from a few major drawbacks: a) Localization Net is optimized for the overall model objective with no direct reference to the amount of distortion to be rectified. b) a model with deeper layers further suffers from the vanishing or exploding gradients, impacting poor reconstruction c) different objects with varied geometric transformations in a scene can reduce prediction confidence of Localization Net. In effect, the incorrect geometric warping by the STN sampler impacts the downstream applications such as object detection, tracking, etc.

\subsection{Novel Regularized Spatial Transformer Networks:} To combat the aforementioned challenges of STN, we perform a self-supervised training \cite{misra2020self} of the Localization Net. Our proposed approach is inspired by deep supervision \cite{lee2015deeply}, allowing better gradient flow in an end-to-end network. First, we introduce an Affine sampler that produces known affine parameters, $\theta_{GT}$ for the arbitrarily selected translation ($t_x,t_y$), rotation ($\phi$), scaling ($S_x,S_y$) values,   
\begin{small}
\begin{equation}
\theta_{GT} = 
\begin{bmatrix}
1 & 0 & t_x\\
0 & 1 & t_y\\
0 & 0 & 1
\end{bmatrix}
\begin{bmatrix}
\cos{\phi} & -\sin{\phi}  & 0\\
\sin{\phi} & \cos{\phi} & 0\\
0 & 0 & 1
\end{bmatrix}
\begin{bmatrix}
S_x & 0 & 0\\
0 & S_y & 0\\
0 & 0 & 1
\end{bmatrix}
\end{equation}
\end{small}
Then, we obtain a sampling grid by warping a regular grid with $\mathcal{T}_{\theta_{GT}}(G)$. An input image or feature map $V_{GT}$ is then bilinear transformed by the new sampling grid, producing deformed $U$. The Localization Net in STN predicts affine parameters, $\theta_p$. Further transforming deformed $U$ with $\mathcal{T}_{\theta_{p}}(G)$, we obtain $V_p$. Ideally, the reverse warped $V_p$ should be equal to $V_{GT}$. Thereby we regularize the Localization Net to predict $\theta_p$ as $\theta_{GT}^{-1}$. The \textsc{RuSTaN} module is shown in Fig. \ref{Fig_Rustan}b and is constrained with two optimization objectives during training,
\begin{small}
\begin{align}
&\mathcal{L}_{\theta} = \frac{1}{N} \sum_{i=1}^{N}\Vert\theta_{GT} \times \theta_{p} - \mathcal{I}\Vert_{1}; \nonumber\\
&\mathcal{L}_{I} = {\Vert V_{GT} -  V_{p} \Vert}_{1};  \nonumber\\
&\mathcal{L}_{\textsc{RuSTaN}} = \lambda_1* \mathcal{L}_{\theta} + \lambda_2* \mathcal{L}_{I}
\end{align}
\end{small}

Where, $\mathcal{L}_{\theta}$ and $\mathcal{L}_{I}$ are the loss components for optimizing affine parameters and images, respectively. $\lambda's$ are the coefficients for the loss components. $N = |\theta_{GT}| = |\theta_{p}|$, where $|.|$ denotes the cardinality operator. $\mathcal{I}$ is the Identity matrix and $\theta_{p}$ is regularized to become $\theta_{GT}^{-1}$ in $\mathcal{L}_{\theta}$. In this experiment, we consider $\mathcal{L}_{\theta}$ and $\mathcal{L}_{I}$ as mean absolute error loss. During inference, the projective distorted images $U$ are directly processed by pre-trained Localization Net of \textsc{RuSTaN} and are expected to produce accurate canonical output, which helps in further downstream applications.

\subsection{Proposed framework} The proposed architecture is shown in Fig. \ref{fig:Archi}, and below, we summarize the individual modules of our architecture.\\
\noindent \textbf{i) Prompt-driven self-supervised learning}: To simulate the impact of degraded weather and learn to nullify the visual deterioration, we introduce climatic diffusion and weather distillation modules, respectively. \\
 \noindent -\textbf{Climatic diffusion}: We leverage a text-to-image diffusion model, ControlNet \cite{zhang2023adding}, for spatial conditioning of climatic variations. Precisely, we control the weather artifacts, i.e., smoke, fog, etc., through a depth map and edge map as control images. The combination is due to the severe weather artifacts increasing with distance from an aircraft while we also need to maintain airport semantic structure. Corresponding weather-conditioned simulated images act as target images during finetuning. Besides, we use customized templates. For example, \textit{``An aircraft is landing in heavy \{*weather\_condition\}}", where \textit{*weather\_condition} can be \{``rain", ``fog", ``smoke", ``haze"\}, etc.\\
\noindent -\textbf{Weather distillation}: Once we obtain a paired correspondence of the degraded weather and a clean landing image, we learn to minimize pixel reconstruction loss. For this purpose, we use Pix2Pix \cite{isola2017image}, a Generative Adversarial Network (GAN)-based image-to-image translation framework. Although any other supervised image translation model can be considered, the advantage of Pix2Pix is the ability to generate realistic, diverse, clear weather images from a relatively small training dataset. Pix2Pix quickly captures the structured noise distribution for individual weather classes and learns to remove them. During real-time landing, the pre-trained Pix2Pix distills weather artifacts to generate clear landing images.\\
\noindent - \textbf{Diffusion-distillation loss}: During training, we stitch disentangled generation of harsh weather with distillation counterpart using self-supervised learning \cite{misra2020self}. Objective becomes,
\begin{small}
    \begin{align}
    &\mathcal{L} = \mathbb{E}_{y,z_0,t,c_t,c_f,\epsilon\sim\mathcal{N}(0,1)}{\Vert y - G(x,\xi{\rho}(z_t,t,c_t,c_f)) \Vert}_{1};  
    \end{align}
\end{small}

Where, y is the target clear day image, and $\xi{\rho}$ is the frozen diffusion network of ControlNet with intermediate noisy image $z_t$ sampled from Gaussian noise $\epsilon\sim\mathcal{N}(0,1)$. At each time step $t$ of climatic diffusion, $\xi{\rho}$ predicts the amount of noise added to $z_t$ given a textual prompt $c_t$, and a conditional control (i.e., depth map or edge map) $c_f$. Leveraging this predicted noise and ControlNet-generated harsh weather image $x$, the Pix2Pix generator $G$ synthesizes clear-day images. The L1 loss (Manhattan distance) between $y$ and $G$ output is used to optimize Pix2Pix GAN components at each time step. Even though without $\xi{\rho}$ prediction, $G$ could map $x$ to $y$, the inclusion of it enables diversity to rectify from various distributions of climatic conditions.\\
\noindent \textbf{ii) Geometric distortion calibration by \textsc{RuSTaN}}: The Pix2Pix generator ($G$) outputs an aircraft landing image free from atmospheric effects, which is then calibrated by the \textsc{RuSTaN} module to obtain canonical representation. The projective transformation warps the input video frame to align the runway parallel to the vertical axis of the image plane. It helps the downstream object detector for reliable localization of the runway elements due to getting pre-trained with ground-truth annotations at no tilting angle of an aircraft which reduces annotation complexity. \\
\noindent \textbf{iii) Runway object detection:} Using YOLO v8 \cite{yolov8}, we detect different runway elements, i.e., centerline, markers, numbers, etc. YOLOv8 outperforms other object detectors in tiny object detection, especially when the aircraft is far from the runway. It helps the pilot perform a vision-based dock while getting reliable warnings and maintain aircraft running over the centerline, thus minimizing any probable wingtip collisions with surrounding airport objects.\\
\noindent \textbf{iv) Inference steps:} First, the aircraft camera data passes through $G$ of the weather distillation module to eliminate any harsh weather visual artifact. Then, it is geometrically calibrated by Localization Net predicted parameters due to landing at varied attitudes. Finally, different runway elements are recognized by object detector.

\section{Experiments and Analysis}
\subsection{Dataset overview}
\begin{figure*}[htbp]
\centerline{\includegraphics[width=\linewidth]{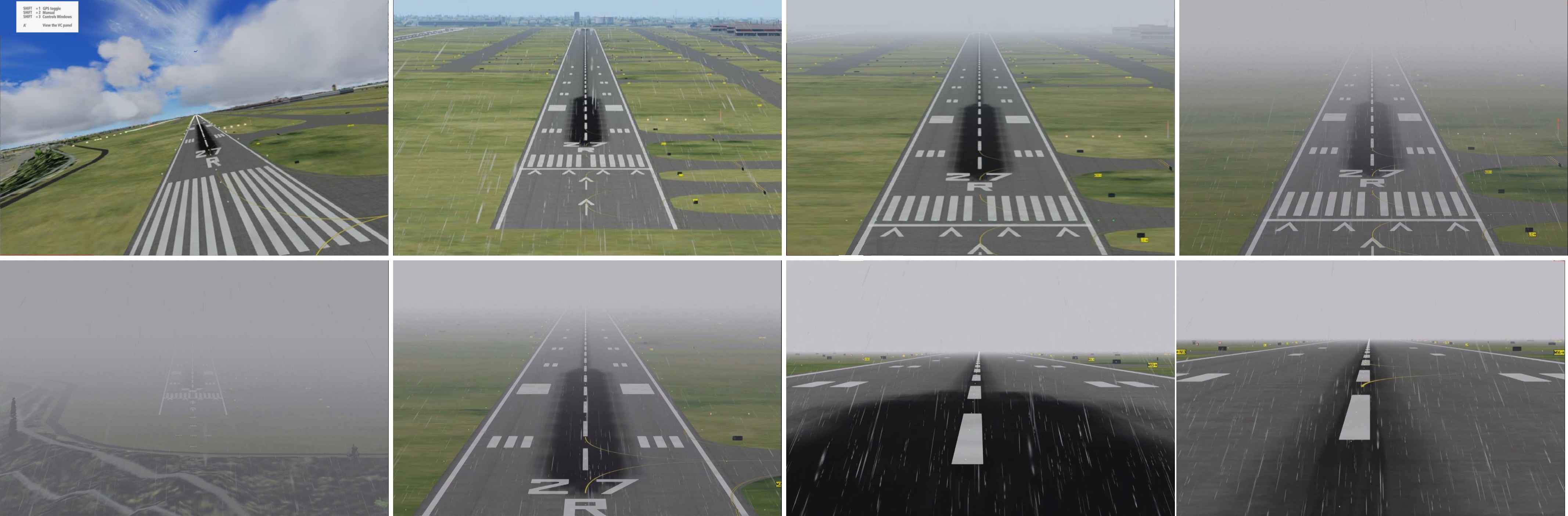}}
\caption{Sample images of the AIRLAD dataset using Lockheed Martin Prepar3D simulator are shown. (Row-1). Landing in a clear day with 3\textdegree attitude, rain, fog, and fog with rain, respectively. (Row-2) shows different instances of landing from decision height to touchdown, featuring the realistic generation of weather artifacts with tire skid marks and crossing taxiway.}
\label{fig:AIRLAD_sample}
\vspace*{5mm}
\centerline{\includegraphics[width=\linewidth]{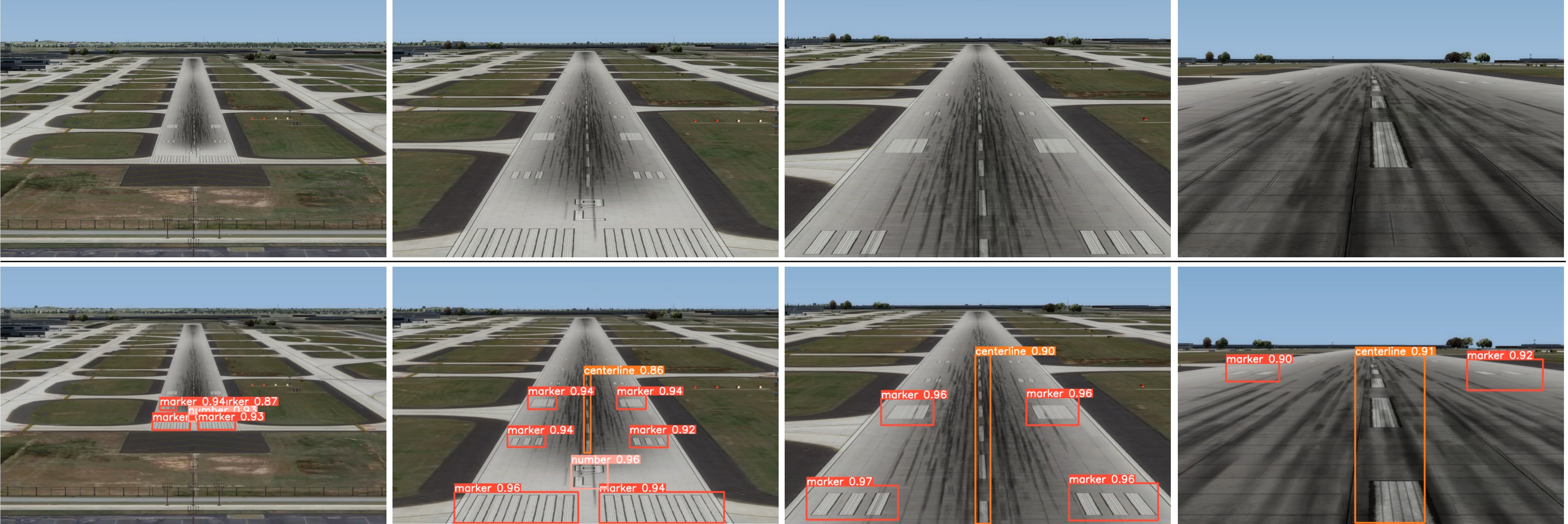}}
\caption{(Row 1) Sample clear-day images of the AIRLAD dataset featuring aircraft landing at KATL airport. (Row 2) Bounding boxes of the detected runway elements by YOLO v8 with corresponding confidence scores are annotated. Thanks to \textsc{RuSTaN} module for effectively projecting images at a vertical axis parallel to the runway even though input images are rotated at 5$^\circ$ to mimic the crosswind landing scenario. Overall, it reduces the high divergence or localization error of the predicted bounding boxes from the actual canonical representation and minimizes the overlap between bounding boxes of nearby runway elements.}
\label{fig:AIRLAD}
\end{figure*}

The vision-based landing plays a pivotal role in providing important visual cues during the critical phases of aircraft landing. Among thousands of airports globally, Hartsfield-Jackson Atlanta International Airport (ICAO: KATL) in the United States consistently secures the busiest airport ranking. ILS or any other subsystem failure in an aircraft significantly impedes landing, thus becoming a bottleneck for traffic management. It motivates us to develop a visual landing scenario for KATL airport, and we term this dataset AIRLAD (AIRport LAnding Dataset). We use a virtual aircraft model, Carenado Commander 114, to simulate the landing scenario using the Prepar3D flight simulation platform. The initial position is three nautical miles from the aiming point with an attitude of 3\textdegree. Presently, the data is captured at 11.00 AM GMT for three different weather phenomenons, namely, rain (RA), fog (FG), and snow (SN). For each phenomenon, the data contains 6,340 video frames at 30 FPS. However, due to the standard simulation parameters, it produces images with limited weather diversity. Further, we synthesize complex weather phenomena for training by incorporating climatic diffusion with textual prompts (i.e., long/short droplets, angle, density variation, cloud, fog mix, etc.). AIRLAD complements a few available runway detection datasets (i.e., LARD \cite{ducoffe2023lard}) but explicitly focuses on landing under weather degradation. Sample AIRLAD images are shown in Fig. \ref{fig:AIRLAD_sample}. 

\subsection{Evaluation metrics}
We use Fréchet inception distance (FID) \cite{heusel2017gans} and inception score (IS) metrics to evaluate the quality of the generated images by the climatic diffusion module. The FID score should be low and IS high to generate diverse, realistic images. Then to access the runway object detection performance, we use Precision (P), Recall (R), F1 score, and Mean Average Precision (mAP). Precision quantifies the proportion of true positives among all positive predictions. Again, recall calculates the proportion of true positives among all actual positives, measuring the model's ability to detect all class instances. The F1 score is the harmonic mean of precision and recall, providing a balanced assessment of a model's performance while considering both false positives and false negatives. mAP is computed by applying a threshold on Intersection over Union (IoU), which measures the overlap between predicted and ground truth bounding boxes, assessing the object localization precision. \text{mAP50} represents mAP at the IoU threshold 0.50, evaluating model accuracy for easy detection. Whereas \text{mAP50-95} averages mAP score across IoU thresholds of 0.50 to 0.95, providing a holistic view of model performance across various detection difficulty levels \cite{Ultralytics}.
\begin{align}
\text{mAP50} &= \frac{1}{N} \sum_{i=1}^{N} \text{AP}_i^{0.5}; \text{mAP50-95} = \frac{1}{N} \sum_{i=1}^{N} \text{AP}_i^{0.5-0.95} \label{mAP50-95}
\end{align}
Where, $AP_{i}^{0.5}$ represents the average precision for class $i$ at IoU threshold 0.5 and $AP_{i}^{0.5-0.95}$ indicates the same for threshold ranging from 0.5 to 0.95. $N$ is the number of classes, which is 3 in our case as we estimate detection performance for three runway classes: marker, centerline, and number.

\subsection{Experimental results}

We train the proposed architecture in Fig. \ref{fig:Archi} for 500 epochs with Adam optimizer of learning rate 1e-5 for three cases: a) without spatial transformation, b) with STN \cite{jaderberg2015spatial}, and c) with \textsc{RuSTaN} and record the test performance in Table \ref{tab_Obj_det}. Test frames are rotated from 0$^{\circ}$ to 15$^{\circ}$ to evaluate runway object detection performance at crosswind landing.

\begin{table*}[!htbp]
\caption{Comparative analysis of the runway object detection performance with and without \textsc{RuSTaN}}
\begin{center}
\renewcommand{\arraystretch}{1.2}   
\setlength{\tabcolsep}{3pt}        
\scalebox{0.9}{
\begin{tabular}{|c||c c c c| c c c c| c c c c|}
\hline
\textbf{Angular}&\multicolumn{4}{c}{\textbf{W/o spatial transformation}} &\multicolumn{4}{|c}{\textbf{With STN \cite{jaderberg2015spatial}}} &\multicolumn{4}{|c|}{\textbf{\shortstack{With \textsc{RuSTaN}}}} \\
\cline{2-13} 
\textbf{deviation} &\textbf{P}&\textbf{R}&\textbf{mAP50}&\textbf{mAP50-95}&\textbf{P}&\textbf{R}&\textbf{mAP50}&\textbf{mAP50-95}&\textbf{P}&\textbf{R}&\textbf{mAP50}&\textbf{mAP50-95}\\
\hline
\hline
0\textdegree
&0.99      &0.97       &0.99       &0.90
&0.99      &0.95       &0.98       &0.92
&\textbf{0.99}      &\textbf{0.98}       &\textbf{0.99}       &\textbf{0.94}\\
\hline
3\textdegree
&0.88      &0.90      &0.91      &0.40
&0.91      &0.92       &0.93       &0.71
&\textbf{0.95}      &\textbf{0.95}       &\textbf{0.96}       &\textbf{0.82}\\
\hline
5\textdegree
&0.70      &0.68       &0.67       &0.21 
&0.78      &0.75       &0.80       &0.65
&\textbf{0.87}      &\textbf{0.84}       &\textbf{0.88}       &\textbf{0.76} \\
\hline
10\textdegree
&0.13      &0.12       &0.10       &0.03
&0.55      &0.58       &0.56       &0.47
&\textbf{0.75}      &\textbf{0.72}       &\textbf{0.79}       &\textbf{0.61} \\
\hline
15\textdegree
&0.08      &0.10       &0.06       &0.02
&0.35      &0.33       &0.37       &0.24
&\textbf{0.41}      &\textbf{0.38}       &\textbf{0.46}       &\textbf{0.32} \\
\hline
\end{tabular}}
\label{tab_Obj_det}
\end{center}
\vspace*{-3mm}
\end{table*}

\begin{figure*}[htbp]
\centerline{\includegraphics[width=0.9\linewidth]{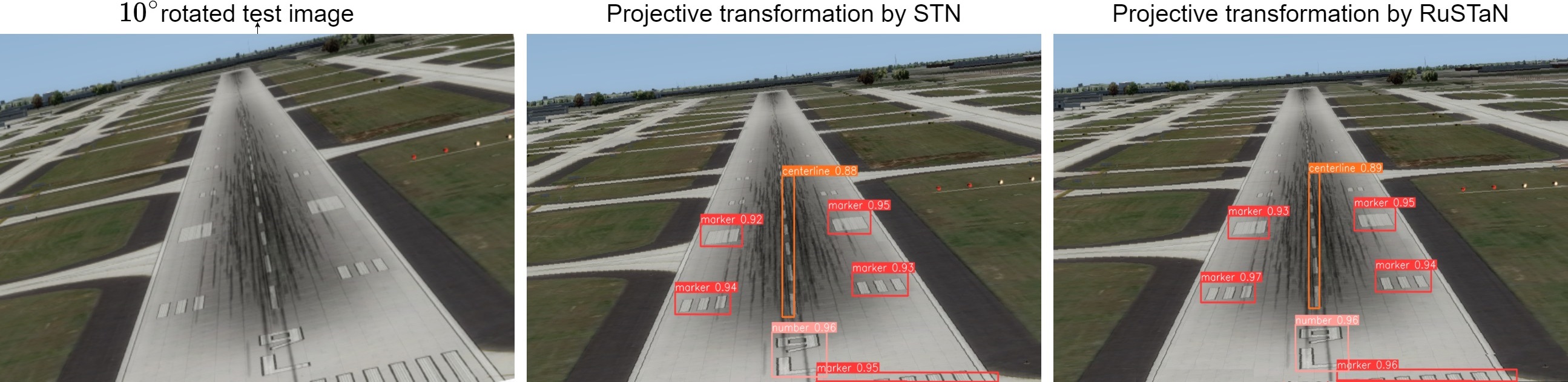}}
\vspace*{-3mm}
\caption{Comparative analysis to restore from projective distortion after rotating test image by 10$^\circ$.}
\label{fig:Calib}
\vspace*{1mm}
\centerline{\includegraphics[width=0.9\linewidth]{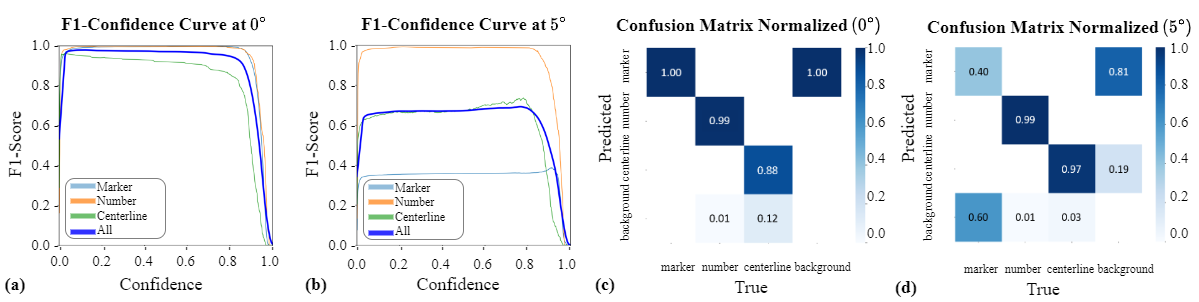}}
\vspace*{-3mm}  
\caption{The class-wise F1 scores in detecting runway objects from the 0$^{\circ}$ and 5$^{\circ}$ rotated test frames are shown in (a) and (b), respectively, to evaluate the crosswind landing performance without using STN or \textsc{RuSTaN}. Corresponding confusion matrices are shown in (c) and (d). Without spatial transformation, the localization error of the object detector steadily increases.}
\label{fig:Conf_F1_Score}
\end{figure*}

Without any spatial transformation, increasing angular variation of test data causes a sharp decline in object detection performance, especially at 15$^{\circ}$ rotation, where none of the runway objects are detected. The Localization Net of STN predicts affine parameters, used to warp an image and restore from projective distortion. Incorporating STN, the detection performance improved significantly due to the warping of rotated test images by the Localization Net predicted affine parameters prior to runway object detection. Effectively, this performance indicates the geometric transformation invariance nature by the learnable Localization Net. Further, utilizing the proposed \textsc{RuSTaN}, the detection performance drastically outperformed STN, particularly for deviations other than 0$^{\circ}$. \textsc{RuSTaN} can quickly calibrate from arbitrary geometric distortions due to leveraging the Localization Net, which is pre-trained by the proposed self-supervised learning approach.  After being calibrated by \textsc{RuSTaN} from the 5$^\circ$ rotated test landing image, the detected runway objects are shown in Fig \ref{fig:AIRLAD} (Row 2). We obtain a more precise runway parallel warped image by \textsc{RuSTaN} in Fig \ref{fig:Calib}, resulting in better runway object detection performance with higher confidence.

The curves in Fig. \ref{fig:Conf_F1_Score} (a, b) illustrate how the F1-score changes as the confidence threshold varies for different angular variations of test images without any spatial transformation. At 0\textdegree, all the classes have a high F1-score, but it drops at 5\textdegree, specifically for the runway marker class having a high aspect ratio in image space. Fig. \ref{fig:Conf_F1_Score} (c, d) indicates a comprehensive evaluation of the model's confusion matrix under the same test scenario. At 0\textdegree, predicted false positives or negatives are minimal. Rotating test data to 5$^{\circ}$ raises the number of false positives, especially for the marker class.  

\begin{table}[htbp]
\caption{Impact of optimizing weather distillation model with diffusion-distillation loss}
\begin{center}
\renewcommand{\arraystretch}{1.2}   
\setlength{\tabcolsep}{3pt}        
\scalebox{0.9}{
\begin{tabular}{|c||c c c c|}
\hline
\textbf{Diffusion-distillation}&\multicolumn{4}{c|}{\textbf{With \textsc{RuSTaN}, (5$^\circ$ rotation) }}\\
\cline{2-5} 
\textbf{loss} &\textbf{P}&\textbf{R}&\textbf{mAP50}&\textbf{mAP50-95}\\
\hline
\hline
Without &0.81      &0.83       &0.79       &0.68\\
\hline
With &\textbf{0.87}      &\textbf{0.84}      &\textbf{0.88}     &\textbf{0.76}\\
\hline
\end{tabular}}
\label{tab_loss}
\end{center}
\vspace*{-3mm}
\end{table}

\begin{figure*}[htbp]
\centerline{\includegraphics[width=\linewidth,height=5.75cm]{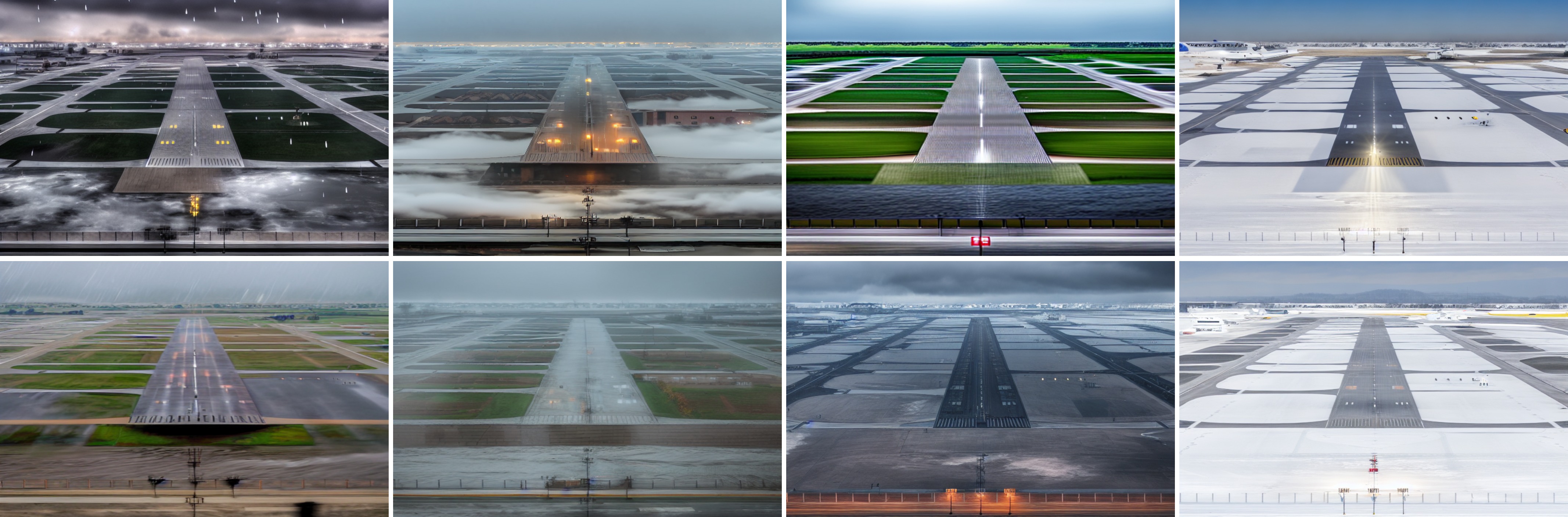}}
\caption{Effect of fine-tuning ControlNet (Row-1) Directly inferencing pre-trained ControlNet with a reference Edge map as a control image produces unrealistic landing images (Row-2). Fine-tuning with harsh weather-conditioned AIRLAD images produces visually plausible synthesized aircraft landing images.}
\label{fig:Diff_Op}
\end{figure*}

\subsection{Ablation analysis}
\noindent \textbf{i. Optimization with diffusion-distillation loss:} The weather distillation module learns to restore clear-day images from the induced visual degradations by the climatic diffusion model. Optimizing the distillation module with the proposed diffusion-distillation loss helps to learn the inverse relationship in the latent space. Thus, we obtain clearer day images, allowing better runway object detection. In Table \ref{tab_loss}, we evaluate the impact of this loss optimization considering the test frames have rotated by 5$^\circ$ for crosswind landing, and they are projective transformed by \textsc{RuSTaN} before object detection. Optimization with diffusion-distillation loss causes approximately 8\% improvement of mAP50-95 metric.

\begin{table}[htbp]
\caption{Evaluation of text-conditioned landing image synthesis on the AIRLAD dataset.}
\vspace*{-3mm}
\begin{center}
\renewcommand{\arraystretch}{1.2}   
\setlength{\tabcolsep}{3pt}        
\scalebox{0.9}{
\begin{tabular}{|c|c  | c |}
\hline
\textbf{Method} &\textbf{FID} $\downarrow$ &\textbf{IS} $\uparrow$\\
\hline
\hline
Pix2Pix \cite{isola2017image} &  15.22 &  21.94   \\
CycleGAN \cite{zhu2017unpaired} & 18.71  &  26.28   \\
ControlNet \cite{zhang2023adding} & 12.06  &  46.29   \\
\hline
\textbf{Ours} &\textbf{8.46}      &\textbf{53.24}    \\
\hline
\end{tabular}}
\label{tab_synthesis}
\end{center}
\vspace*{-3mm}
\end{table}
\noindent \textbf{ii. Impact of fine-tuning with simulated weather degraded images:} In Fig. \ref{fig:Diff_Op}, we compare the result with and without fine-tuning ControlNet \cite{zhang2023adding}. We observe that lack of airport context results in visually implausible landing image synthesis in (Row 1) as these images contain the abrupt spread of fog artifacts, building structures just beside the runway, multiple vertical lines on the runway, light source on the markers, etc. On the contrary, fine-tuning ControlNet with simulated weather-degraded images from our AIRLAD dataset as target images causes a realistic generation of aerial landing images under harsh climatic conditions in (Row-2).

\noindent \textbf{iii. Performance analysis of diffusion-based weather synthesis:} We compare the performance of the diffusion model against other image generation models in generating weather-degraded images. Notably, Pix2Pix \cite{isola2017image} generates images on paired data, while CycleGAN \cite{zhu2017unpaired} generates images on unpaired data. Both these models perform image-to-image generation without any additional text prompt as input. They are trained with AIRLAD images of different weather phenomenon categories. On the other hand, ControlNet \cite{zhang2023adding} takes additional text prompts but generates images without fine-tuning by diffusion-distillation loss. Finally, we observe fine-tuning ControlNet with proposed diffusion-distillation loss significantly improved FID score by 3.6 margin and produced images have better sample quality with content diversity due to better capturing the whole image distribution \cite{isola2017image, zhu2017unpaired} .

\begin{figure}[htbp]
\centerline{\includegraphics[width=\linewidth, height=4.5cm]{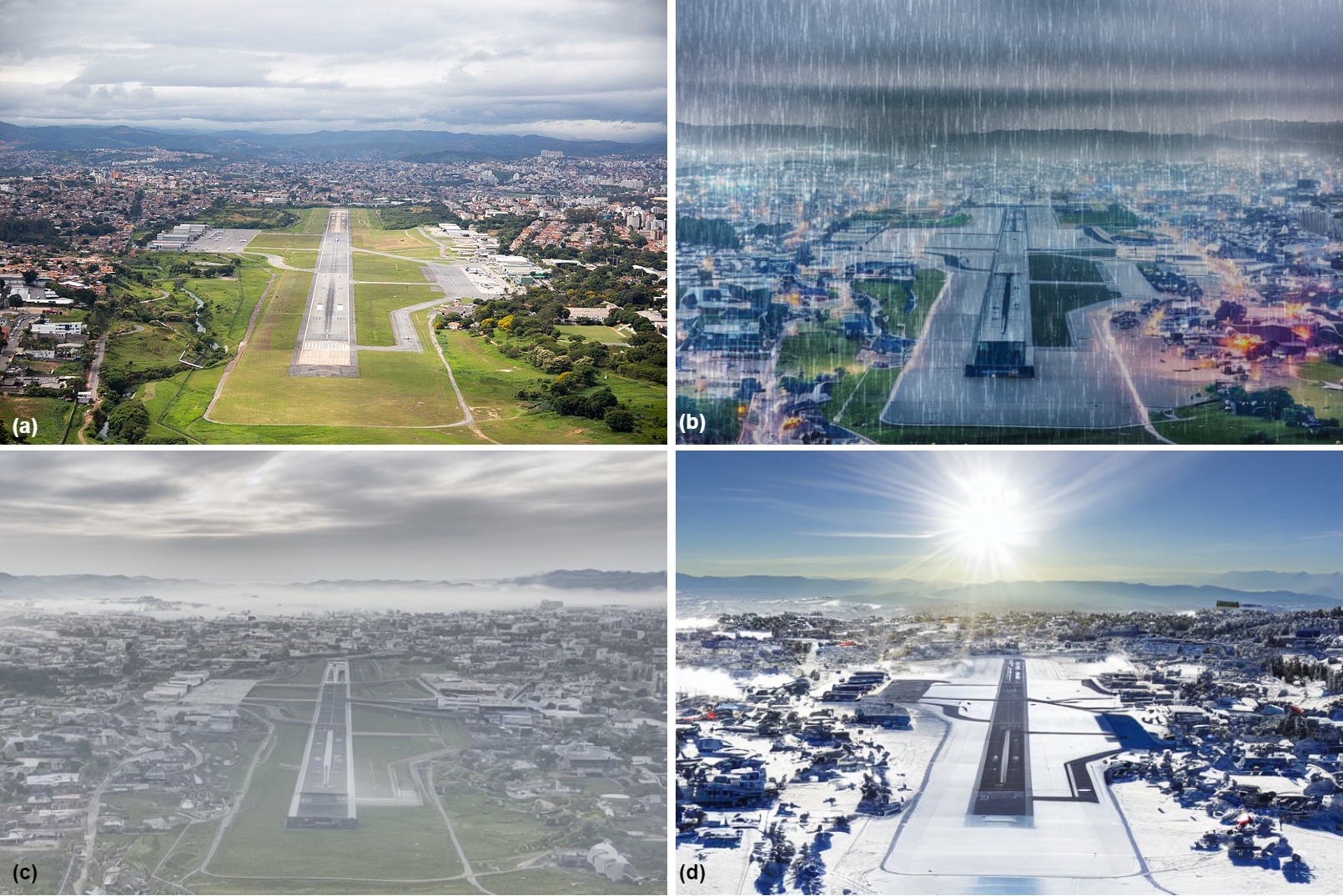}}
\vspace*{-2mm}
\caption{(a) Ground truth clear day landing image `Aeroporto da Pampulha 6.jpg' \cite{Aeroporto} and climatic diffusion module generated images with (b) rain (c) fog (d) snow weather conditions.} 
\vspace*{-2mm}
\label{fig:Real_weather}
\end{figure}

\noindent \textbf{iv. Evaluation on real aerial landing images:} To validate that our simulated data agrees with the real harsh weather distribution, we seek to synthesize weather artifacts on an actual aircraft landing image, `Aeroporto da Pampulha 6.jpg' \cite{Aeroporto}, acquired over Belo Horizonte, Brazil, in April 2014. We observe that the images generated (Fig. \ref{fig:Real_weather}) by the climatic diffusion module capture the weather degradation scenarios well while maintaining consistency with the base image in terms of overall runway structure.

\section{Conclusion}
This paper proposes a prompt-based visual weather degradation synthesis method to form a landing scenario paired dataset under clear-day and harsh weather conditions. Thereby, a novel diffusion-distillation loss helps optimize the weather distillation network to learn to remove the harsh weather visual artifacts. Further, directly optimizing the Localization Net of \textsc{RuSTaN} for the affine sampler predicted parameters causes to perform better runway object detection at crosswind landing. Experimental results over the proposed AIRLAD dataset and ablation analysis prove our real-time situational aware runway object detection capability for vision-based aircraft landing. In the future, we will apply optical character recognition to the detected runway number class images. Also, we plan to enlarge our dataset with real and simulation-based landing scenarios at other airports and evaluate the model performance.


\clearpage


\end{document}